\documentclass{article}
\usepackage{ijcai21}
\usepackage{times}
\usepackage{soul}
\usepackage{url}
\usepackage[hidelinks]{hyperref}
\usepackage[utf8]{inputenc}
\usepackage[small]{caption}
\usepackage{graphicx}
\usepackage{amsmath}
\usepackage{booktabs}
\usepackage{epsfig}
\usepackage{enumitem}
\usepackage{color}
\usepackage{subfigure}
\usepackage{setspace}
\usepackage{multirow}
\usepackage{multicol}
\usepackage{amsfonts}
\usepackage[switch]{lineno}
\title{Enhance Convolutional Neural Networks with Noise Incentive Block}

\author{
Menghan Xia$^1$\and
Yi Wang$^1$\and
Han Chu$^{2}$ \and
Tien-Tsin Wong$^{1}$
\affiliations
$^1$Department of Computer Science and Engineering, CUHK\\
$^2$Guangdong Provincial People’s Hospital, Guangdong Academy of Medical Sciences\\
\emails
\{mhxia, yiwang, ttwong\}@cse.cuhk.edu.hk,
zq1992@gmail.com
}

\begin{document}

\maketitle

\begin{abstract}
As a generic modeling tool, Convolutional Neural Networks (CNNs) have been widely employed in image generation and translation tasks. However, when fed with a flat input, current CNN models may fail to generate vivid results due to the spatially shared convolution kernels. We call it the flatness degradation of CNNs. Unfortunately, such degradation is the greatest obstacles to generate a spatially-variant output from a flat input, which has been barely discussed in the previous literature. To tackle this problem, we propose a model agnostic solution, i.e. Noise Incentive Block (NIB), which serves as a generic plug-in for any CNN generation model. The key idea is to break the flat input condition while keeping the intactness of the original information. Specifically, the NIB perturbs the input data symmetrically with a noise map and reassembles them in the feature domain as driven by the objective function. Extensive experiments show that existing CNN models equipped with NIB survive from the flatness degradation and are able to generate visually better results with richer details in some specific image generation tasks given flat inputs, e.g. semantic image synthesis, data-hidden image generation, and deep neural dithering.
\end{abstract}

\section{Introduction}
\label{sec:intro}

Recently, Convolutional Neural Networks (CNNs) have demonstrated great success in various image processing and computer vision applications~\cite{Simonyan2015,ParkCVPR19}. They can be roughly categorized into two main directions, image understanding and image generation. Image understanding usually summarizes massive information into a more abstract form, like semantic segmentation. Just the opposite, Image generation aims to create massive information from an abstract image, such as semantic image synthesis. However, existing image generation models usually face a vital but easily overlooked problem. When CNNs are fed with flat inputs, they may fail to generate vivid results.
%
\begin{figure}[!t]
	\centering
	\includegraphics[width=1.0\linewidth]{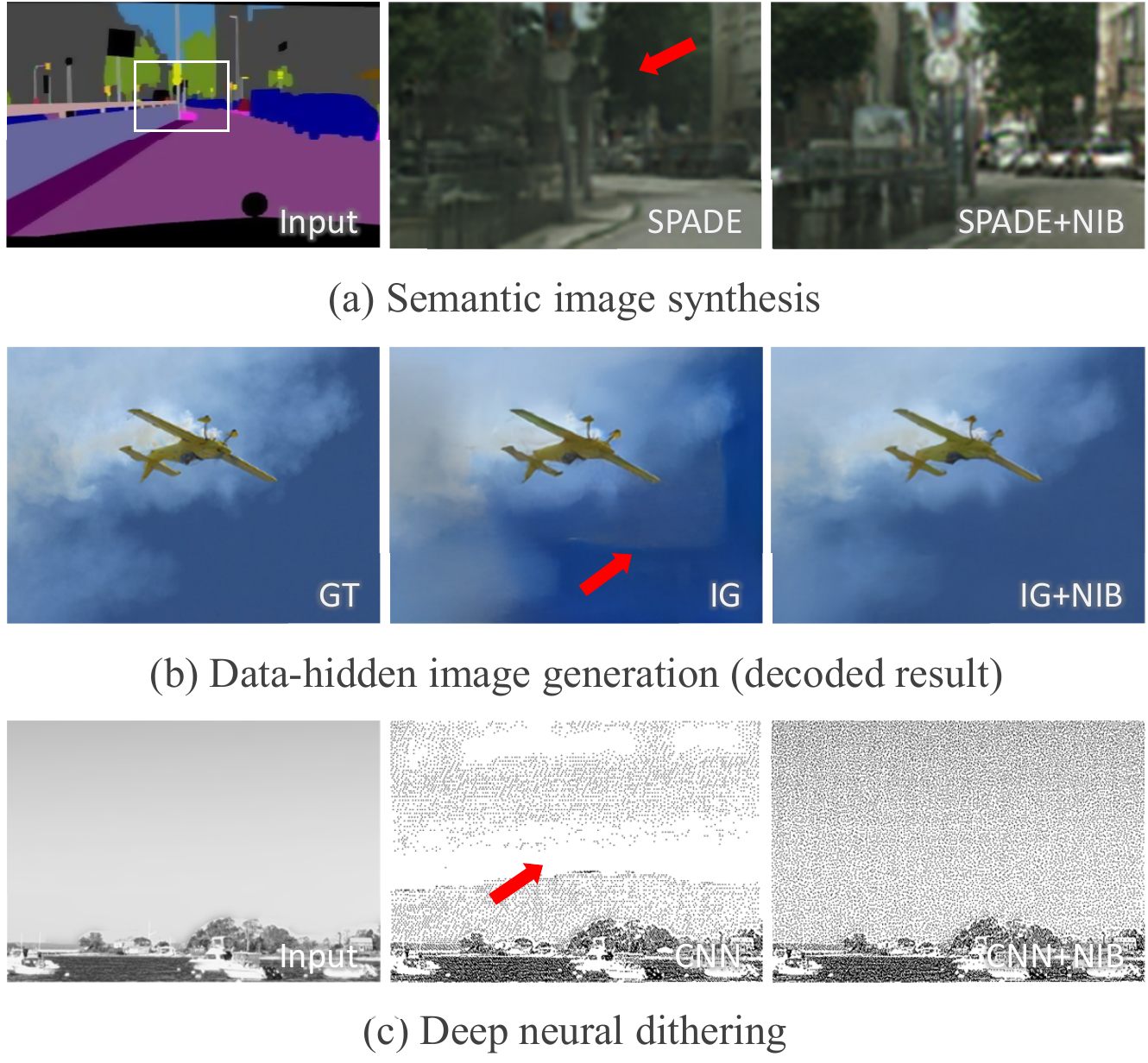}
	\caption{Flatness degradation examples: (a) SPADE~\protect\shortcite{ParkCVPR19} struggles to generate natural details in large homogeneous-label regions. (b) Invertible grayscale (IG)~\protect\shortcite{XiaL2018} fails to accurately encode color information into the grayscale image in flat regions. (c) Typical CNNs fail to dither constant-grayness regions. Notably, our proposed Noise Incentive Block (NIB) addresses these problems effectively.}
 	\label{fig:intro} 
\end{figure}

It is the nature inherited from the convolution operation with spatially shared kernels, i.e. given a constant function $f(x) \equiv c$ and an arbitrary local kernel function $k(x)$, the convolution of $f(x)$ and $k(x)$ can only produce another constant function $g(x) \equiv c\cdot\mu(k(y))$, where $\mu(\cdot)$ computes the mean value. Regarding this, a standard CNN model, integrating cascaded convolutional layers with bias terms and activation functions, retains this property as well. That is, given a flat input $\mathbf{X}$, the transformation by CNN degrades to a scaling operation $\mathbf{Y}=\alpha \mathbf{X}$, where the scalar $\alpha$ is determined by the CNN parameters and $\mathbf{X}$, and hence the output $\mathbf{Y}$ is flat for sure. We call this phenomenon~\textit{flatness degradation} throughout this paper.
Theoretically, it will always happen when the receptive field of a CNN (or its submodules) is filled with flat data, which is inevitable with sparse input. Figure~\ref{fig:intro} demonstrates how flatness degradation affects several typical applications: (a) Semantic image synthesis that synthesizes photo-realistic textures from piece-wise flat semantic layout; (b) Data-hidden image generation that encodes extra information through imperceptible intensity variation; (c) Neural dithering that reproduces tone through discrete halftone patterns. 

\begin{figure}[!t]
	\centering
	\includegraphics[width=1.0\linewidth]{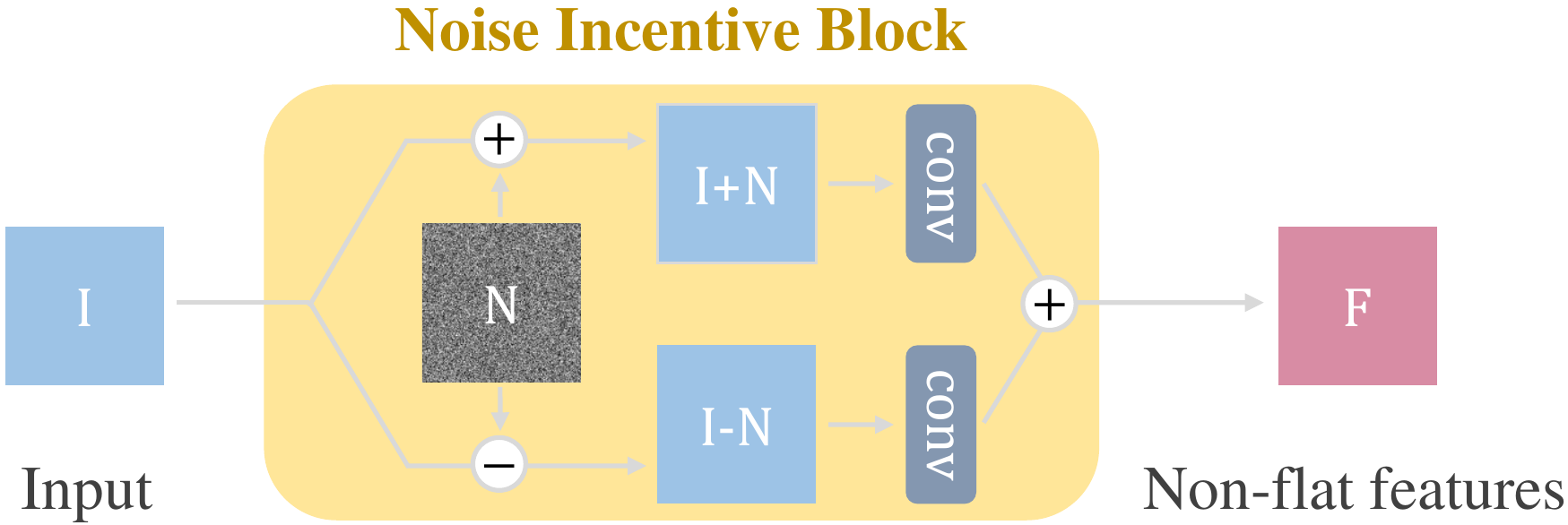}
	\caption{Diagram of the Noise Incentive Block (NIB). The NIB transforms the input data $\mathbf{I}$ into locally non-flat features, on which CNNs are potentially free of flatness degradation issue. The symmetric noise addition guarantees the information restorability nicely.}
 	\label{fig:noise_cnn}
\end{figure}

To circumvent the flatness degradation of CNNs, we propose a generic plug-in block, Noise Incentive Block (NIB). Basically, it is based on a straightforward insight that by reforming the input data to be locally non-flat, the CNN would never work on flat signals.  More importantly, our proposed reform never compromise the input information fidelity. In particular, the NIB perturbs the input data symmetrically with a noise map and then reassemble the two complementary noisy variants in the feature domain (see Figure~\ref{fig:noise_cnn}). The design rationale is that the noisy perturbance breaks flat input condition while the symmetric perturbance strategy generates two complementary components that reserves the information intactness together. In other words, the NIB enables CNNs to generate adaptive output free of flatness degradation, and meanwhile guarantees the information intactness for further task-oriented usage.
This two important functionalities of NIB are verified and studied through various experiments. 
Results tell that equipping CNN models with NIB avoids flatness degradation completely, which even comes with performance gain in general cases. Such advantages are further confirmed by the evaluations on related image generation applications.

Note that, our proposed NIB is a generic solution to tackle flatness degradation issue of CNNs, however, it is possible to have more sophisticated alternatives in specific application scenarios. In some sense, this work serves as a discussion opener and expects to inspire further studies on this problem.
To summarize, this work makes the contributions:
\setitemize[1]{itemsep=1pt,partopsep=0pt,parsep=\parskip,topsep=3pt}
\begin{itemize}
	\item We raise the concept of flatness degradation for the first time, which is demonstrated to degrade the performance of CNN-based generative models.
	\item We propose a simple but effective convolutional plug-in, Noise Incentive Block that resolves flatness degradation for CNNs in a generic manner.
	\item Our proposed method consistently improves the performance of multiple existing state-of-the-art models in relevant applications.
\end{itemize}

\section{Related Works}
\label{sec:background}

We briefly review existing noise utilization in deep neural networks (DNNs), and then discuss with recent works on spatially variant convolution designs.

\paragraph{Promoting training efficiency.}
It is a non-convex optimization problem to train DNNs and there are usually many saddle points and local optima that are associated with bad generalization~\cite{Gerardo2015}. Anyhow, despite the highly complex landscape, Stochastic Gradient Descent (SGD)~\cite{Leon17} and its variants are practically successful in training DNNs. Empirical evidences~\cite{ZhangICLR17,KeskarICLR17} imply that the noise (or say randomness) in Stochastic Gradient Descent (SGD) is very crucial and enables SGD to achieve good optima. The theoretic insights behind these phenomena are discussed in~\cite{ZhouICML2019}.
For data augmentation, injecting noise to input data is leveraged to improve the model robustness to adversarial noise~\cite{Jin15}. Also,~\cite{Neelakantan2015} finds that adding noise to gradient potentially benefits the network training. Noh et al.~\cite{NohNIPS17} propose that the noise injection to the DNN neurons actually optimize the lower bound of the true objective, and a tighter lower bound can be achieved by carefully constructing multiple samples per training examples. 
Different from the above methods that take dynamic noises to introduce randomness to the training, our model just utilizes the local non-flatness nature of the noise map to break flatness, so even a stationary noise map is feasible in our case.

\paragraph{Modeling generative space.}
Generative models generally construct the generation space by exploiting random noises under prior distributions (e.g. normal distribution), so that the target domain could be sampled by controlling the input variables. As the most popular generative framework, Generative adversarial networks (GAN)~\cite{GoodfellowNIPS14} and Variable Autoencoder (VAE)~\cite{Kingma19} model the generation space from the input variables and from the latent variables respectively. To allow user control of the generation, conditionally generative models~\cite{ZhuNIPS17,IsolaCVPR17} synthesize the target domain samples from a preset random distribution and the conditional input. In particular, the conditional input can be either a visually perceptible image~\cite{ParkCVPR19} or a latent variable from another distribution function~\cite{KarrasCVPR19}.
Seemingly, conditional GANs taking as input a conditional image and a noise map, is similar to our noise incentive block. In essence, there is a significant difference. Conditional GANs require the noise map to be dynamically sampled under a well-defined distribution function, by which the noise map carries distribution information to the GAN models for output attributes modeling. In contrast, the noise map equipped in our NIB just serves as a medium of local non-flatness, which offers no information and never affects the property of the output.

\paragraph{Content-adaptive convolution.}
Some recent researches~\cite{Verma17,Dai17,SuH19} raise concerns about the convolutional flexibility under spatial sharing scheme. Particularly, it may be sub-optimal to have the same convolutional filter banks applied to all the images and all the pixels irrespective of their content. Dynamic filters~\cite{Verma17,Dai17,Hux19} and pixel-adaptive convolution~\cite{SuH19} are two well-known techniques to alleviate this restriction, both of which share the same spirit of dynamically generating spatially varying convolution kernels or weighting coefficients from the input features. Indeed, these content-adaptive convolution modules boost performance clearly in mainstream tasks. However, they still suffer from flatness degradation since their spatially varying kernels are generated from the input features by typical convolutions. In this regard, our proposed noise incentive block can be applied to enhance these techniques.

\section{Noise Incentive Block}
\label{sec:noise_paradigm}

To address the flatness degradation for CNNs, we propose a generic plug-in, Noise Incentive Block (NIB), which can be used along with standard convolution layers. The conceptual diagram is illustrated in Figure~\ref{fig:noise_cnn}.

\subsection{Main Idea}
\label{subsec:idea}

Since flatness degradation happens upon flat input, an intuitive idea is to represent the input data in a spatially variant fashion. To this end, we propose to construct a convolutional block that perturbs the input data in a spatially varying manner and preserves the input information from contamination. Below, we discuss how the two characteristics are achieved by our proposed NIB.

\paragraph{Spatially variant proxy.}
We propose to perturbs the input data with a spatially variant proxy. Formally, this proxy is required to satisfy three conditions: (i) non-zero gradient, to avoid flat patch for any cross-pixel convolutional kernels. (ii) non-periodic repeating unit, to avoid the restriction of generating regularly repeated values (see "Regular grid" in Fig~\ref{fig:noise_mode_comp}(b)). (iii) independence on the input data, to avoid information contamination.
Naturally, the random noise map, with independent and identically distributed elements, is a qualified choice. Firstly, since each element is sampled as an independent random variable, the noise map has non-zero gradient almost everywhere and is free of periodic repeating values. Besides, the noise map is generated in a purely random fashion, which definitely has no correlation with the input data. Therefore, the random noise map is a good match to our requirements.
According to Figure~\ref{fig:noise_mode_comp}, the distribution function used for sampling the noise map makes little difference. As default, we take a stationary white noise map $\mathbf{N}$ sampled from $\mathcal{N}(0.0, 0.3)$.

\paragraph{Introduce mechanism.}
With the spatially variant proxy, e.g. the noise map $\mathbf{N}$, we propose to perturbs the input data $\mathbf{I}$ in a symmetric manner, i.e. $\mathbf{I}+\mathbf{N}$ and $\mathbf{I}-\mathbf{N}$, and then resemble them in an adaptively learned feature domain to obtain the potentially non-flat feature representation $\mathbf{\tilde{I}}=f_1(\mathbf{I}+\mathbf{N}) + f_2(\mathbf{I}-\mathbf{N})$. As a default setting, $f_1$ and $f_2$ are implemented with two convolutional layers. These operations constitute our proposed Noise Incentive Block (NIB), as shown in Figure~\ref{fig:noise_cnn}. By this way, the input data $\mathbf{I}$ is represented by $\mathbf{\tilde{I}}$ that is endowed with the potential of offering the intact information of $\mathbf{I}$ and getting rid of flatness degradation.
It worth noting that the two functionality of NIB are not hard properties but potentials to be realized by the adaptively learned $f_1$ and $f_2$ under task-oriented loss functions.

Despite flatness degradation risking every convolution layer, it is unnecessary to substitute all convolutional layers with our NIBs. In fact, the degradation-free potential of NIBs can be transferred to its successor layers. Specifically, supposing the precursor layer is free of flatness degradation, it could generate a non-flat feature maps such that its successor layer inherit this potential accordingly. In other words, adopting a NIB as the first layer could potentially secure the whole CNN model free of flatness degradation.

%
\begin{figure}[!t]
	\centering
	\includegraphics[width=1.0\linewidth]{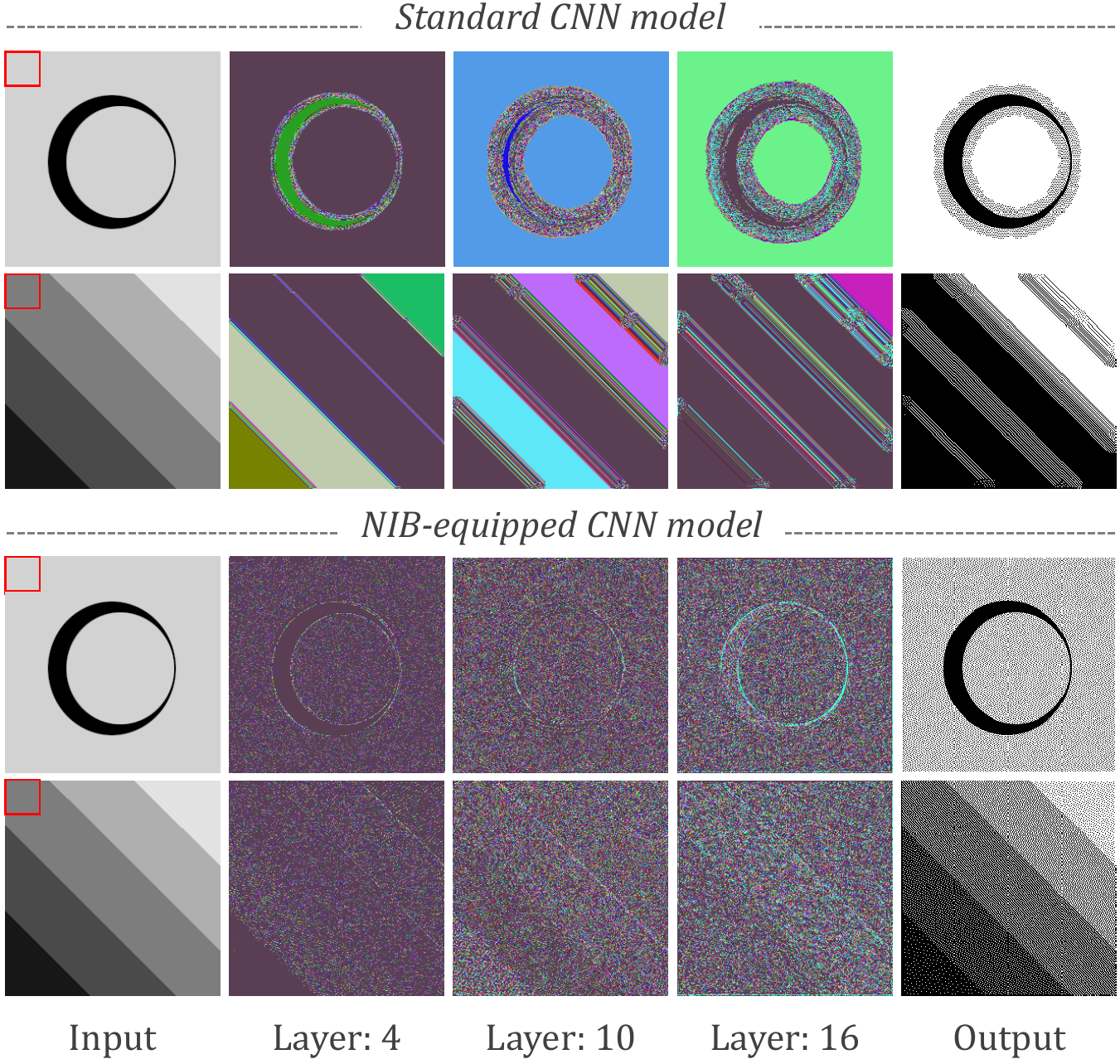}
	\caption{CNN based halftoning: standard model (a) vs. NIB-equipped model (b). The first feature channel from layer 4, 10, 16 are linearly rescaled and visualized as color-coded maps. Both models share the same network architecture (i.e. ResNet), whose receptive field size ($41 \times 41$) is marked on the $256 \times 256$ input images.}
	\label{fig:activation}\vspace{-0.5em}
\end{figure}

\subsection{Model Analysis}
\label{subsec:analysis}

We equip typical CNNs with the proposed NIB to verify its functionalities and explore the working mechanisms. In particular, the questions below are investigated sequentially.

\paragraph{How does NIB work?}
For clear visualization purpose, we explore this problem by applying CNNs for image halftoning~\cite{Ulichney1987}. Formally, given a grayscale image $\mathbf{I}_g$, the CNN model is required to generate a halftone image $\mathbf{I}_h$ that reproduces the tone of $\mathbf{I}_g$ with binary pixels. For simplicity, we only consider the tone similarity for quality assessment, though other properties (e.g. structure similarity, blue noise profile, etc.) are also necessary to quality halftones. The loss function and training details are provided in the supplementary materials.
We take a ResNet consisting of 10 residual blocks~\cite{HeCVPR16} as the baseline model, and the NIB equipped model has a NIB plugged in the first layer. Figure~\ref{fig:activation} compares the results on two typical example. Due to flatness degradation, the standard model can only generate halftone pattern in the surroundings of structure edges, where the radial neighborhood gets larger as the layers get deeper. This indicates that enlarging the receptive field of the CNN can only alleviate the convolution degradation for those deeper layers while the shallower layer still get stuck in those locally flat regions. In contrast, our proposed NIB-equipped model is free of this restriction and can synthesize non-flat feature maps adaptively at all convolution layers.

\begin{table}[!t]
	\centering
	\small
	\renewcommand{\tabcolsep}{4.5pt}
	\renewcommand\arraystretch{1.2}
	\caption{Quantitative evaluation on CNN based halftoning. Higher PSNR means better quality.}\vspace{-0.5em}
	\begin{tabular}{ccccc}
		\hline
		\multirow{2}{*}{Dataset}  & \multicolumn{2}{c}{ResNet}               & \multicolumn{2}{c}{U-Net}    \\
	    \cline{2-5}               & Standard         & NIB-equipped           & Standard      & NIB-equipped         \\ \hline
	    \textit{COMIC}            & $35.837$	        & $ \textbf{37.709}$     & $34.510$      & $\textbf{37.063}$	 \\
		\textit{VOC}              & $36.051$         & $\textbf{36.461}$      & $34.764$      & $\textbf{35.971}$     \\ \hline
	\end{tabular}
	\label{tab:ablation_study}\vspace{-0.5em}
\end{table}

To demonstrate the advantages in general cases, we further quantitatively evaluate the models on comic pictures and natural images. Specifically, $93$ comic frames of resolution $994 \times 1500$ (about 6 pictures per frame) are collected from the Internet, named \textit{COMIC}; $3367$ images are collected from VOC2012 dataset~\cite{VOC2012} and resized to $256 \times 256$, named \textit{VOC}. Two kinds of CNN architecture with different receptive fields: ResNet ($41 \times 41$) and U-Net ($183 \times 183$), are employed for comparison. Table~\ref{tab:ablation_study} tabulates the results. For both CNN architectures, our NIB-equipped model outperform the standard model on \textit{COMIC} clearly since comic pictures almost are filled with flat patches. Interestingly, such superiority is even maintained on the natural image dataset \textit{VOC} when natural images barely contains constant patches.
It is probably because that the CNNs' shallower layers gets stuck in those locally flat spot scattered in smooth regions and thus hinders the model's capability to some extent. Moreover, as the flat area increases, more layers get affected by flatness degradation. Anyhow, NIB-equipped model has no such issue.

\begin{figure}[!t]
	\centering
	\includegraphics[width=1.0\linewidth]{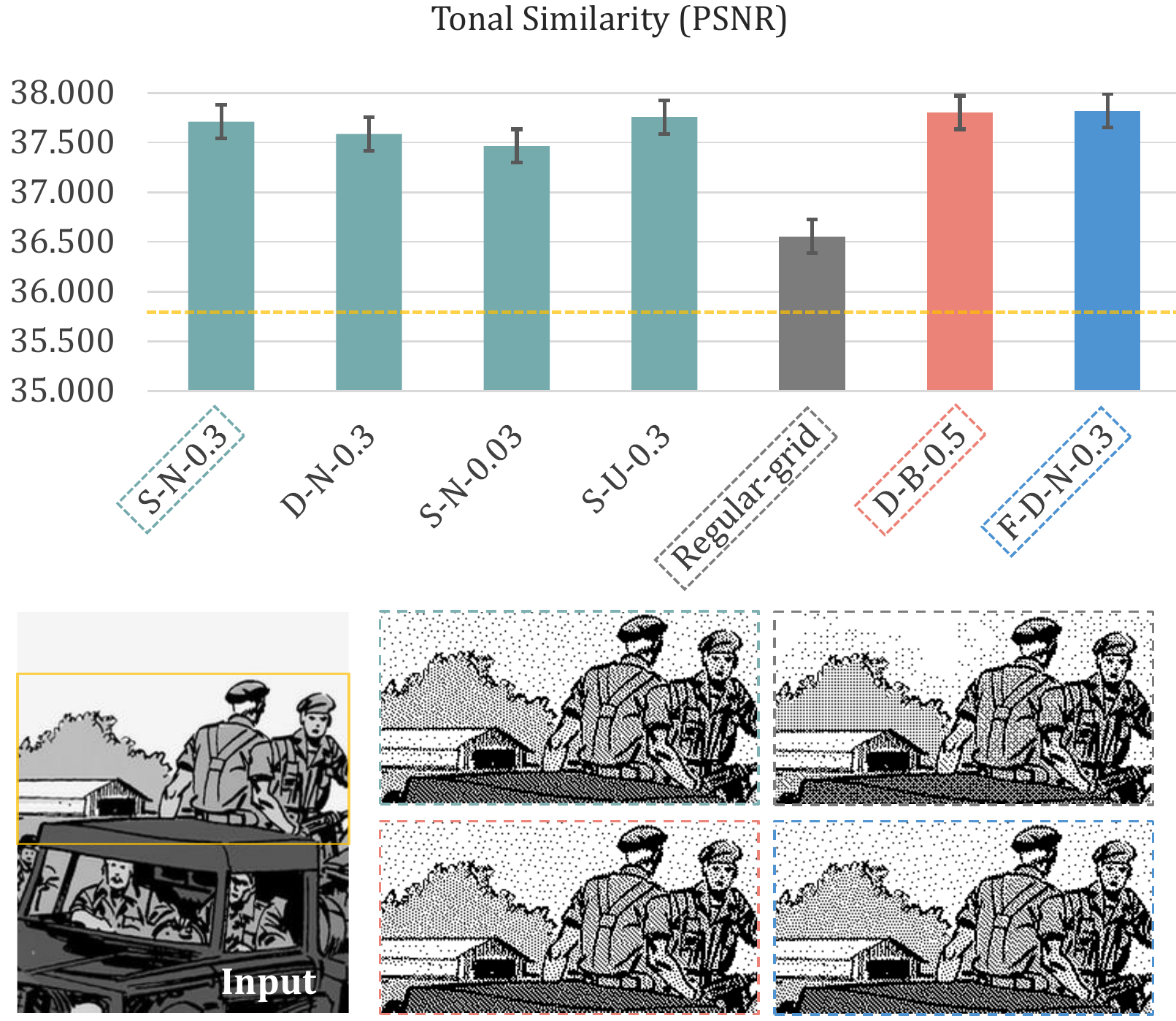}
	\caption{NIB variant comparison based on different noise generation modes. The yellow dash line denotes the model accuracy without NIB applied. The short lines above the bars exhibit the standard deviations. As marked with color dash boxes, four typical variants have their qualitative results compared in the bottom row.}
	\label{fig:noise_mode_comp}\vspace{-0.5em}
\end{figure}

\paragraph{What role is the noise map?}
As discussed in Section~\ref{subsec:idea}, the noise map only serves as a spatially variant proxy that is passively exploited by the task-driven NIB to build non-flat feature maps. In some sense, it is a kind of incentive that enables convolutions to work fluently even the input data is a singular constant map. In this regard, the noise map used in NIB could be either dynamic or stationary and generated from whatever distribution. 
To verify these hypotheses, we compare the NIB variants that adopt different noise injection modes. Specifically, we take the Stationary noise generation from Normal distribution $\mathcal{N}(0.0,0.3)$ as the reference group (\textbf{S-N-0.3}), and further construct four control groups: dynamic noise from $\mathcal{N}(0.0,0.3)$ (\textbf{D-N-0.3}), stationary noise from $\mathcal{N}(0.0,0.03)$ (\textbf{S-N-0.03}), and stationary noise from uniform distribution $\mathcal{U}(-0.3,0.3)$ (\textbf{S-U-0.3}). Here, dynamic noise means freshly generating a noise map for each input and stationary noise means using a fixed noise map for all input data. In addition, a regular grid with alternative black and white lines, is compared as an baseline of disobeying the principle of non-periodic repeating unit. In this study, the ResNet is employed and the quantitative evaluation is performed on \textit{COMIC} dataset.
Figure~\ref{fig:noise_mode_comp} compares the performance of the NIB variants that take different noise injection modes.
Except for the regular grid case, all variants achieve comparable performance and address flatness degradation effectively (see the constant-valued "sky" and "tree" regions). It means that the functionality of NIB is insensitive to the noise type and injection mode. This justifies our hypothesis that the noise map offers no information but passively serves as certain incentive medium. Notably, the periodicity of the regular grid is absorbed by the model during training, as the halftone patterns illustrates. This extra constraint inevitably impedes the model flexibility and thus degrades the performance.

In practice, our proposed NIB, formulated as $\mathbf{\tilde{I}}=f_1(\mathbf{I}+\mathbf{N})+f_2(\mathbf{I}-\mathbf{N})$, can be naturally extended to other paradigms under the same concept. For instance, we may introduce the perturbance by an impulse noise map $\mathbf{B}$ with each element sampled from Bernoulli distribution, and then it turns into $\mathbf{\tilde{I}}=f_1(\mathbf{I}\cdot\mathbf{B})+f_2(\mathbf{I}\cdot(1-\mathbf{B}))$. Also, the perturbance even could be performed in the feature domain only, i.e. $\mathbf{\tilde{I}}=f_1(\mathbf{I})+f_2(\mathbf{N})$, which is especially suitable to the case of semantic input.
Accordingly, they are evaluated in Figure~\ref{fig:noise_mode_comp} as two variants of new paradigms (\textbf{D-B-0.5} and \textbf{F-D-N-0.3}), and present equally decent performance.

\paragraph{Does NIB contaminate data?}
As one of the required features, the NIB is expected to reserve the intactness of the input information. To verify this, we employ NIB into an autoencoder model that reconstructs the input from a learned latent representation. For comparison, we build a naive baseline that adds white noise to the input data.
The rationale is that if the NIB contaminates the input data, the reconstruction accuracy will be affected consequently. The detailed experiment setting is provided in the supplementary material. Figure~\ref{fig:proxy_acc} illustrates an comparative example. Measured by PSNR, the quantitative evaluation on a dataset (3367 images) gives: \underline{47.95 db} (Autoencoder), \underline{35.54 db} (Autoencoder+noise), \underline{48.56 db} (Autoencoder+NIB), respectively. It shows that our proposed NIB causes no data contamination with respective to normal convolution layers, and even gains improvements due to the efficiency of no flatness degradation.

\begin{figure}[!t]
	\centering
	\includegraphics[width=1.0\linewidth]{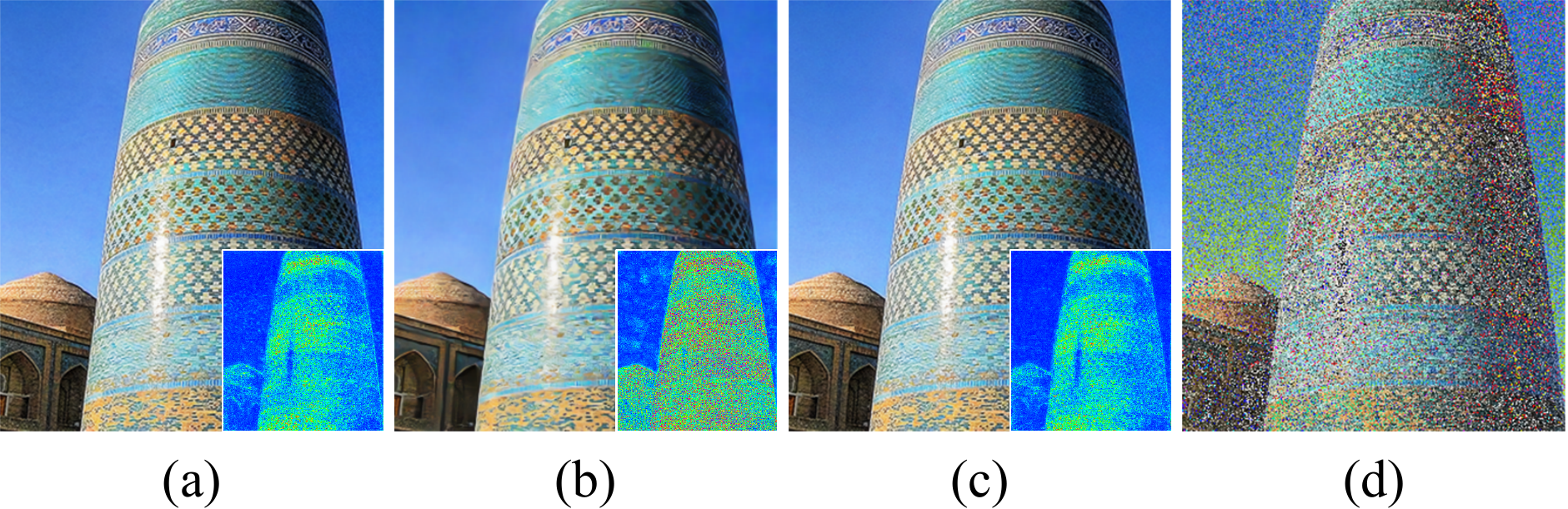}
	\caption{Reconstruction accuracy of: Autoencoder (a), additive noise imposed Autoencoder (b), and NIB equipped autoencoder (c). The color-coded error maps are attached respectively. For reference, the noisy image is shown in (d). Thanks to the symmetric noise mechanism, NIB makes no harm to information reconstruction, while additive noise ruins the fine structures irrevocably.}
 	\label{fig:proxy_acc}
\end{figure}

\section{Application Evaluation}
\label{sec:experiment}

We evaluate our proposed Noise Incentive Block (NIB) on three applications, i.e. semantic image synthesis, data-hidden image generation and deep neural dithering, all of which are required to generate spatial variation from potential flat input. Typically, semantic image synthesis utilizes CNNs with global receptive field for context interpretation, while data-hidden image generation and deep neural dithering adopt CNNs with local receptive field for feature translation.

\subsection{Semantic Image Synthesis}
\label{subsec:semantic_synthesis}

\begin{figure}[!t]
	\centering
	\includegraphics[width=1.0\linewidth]{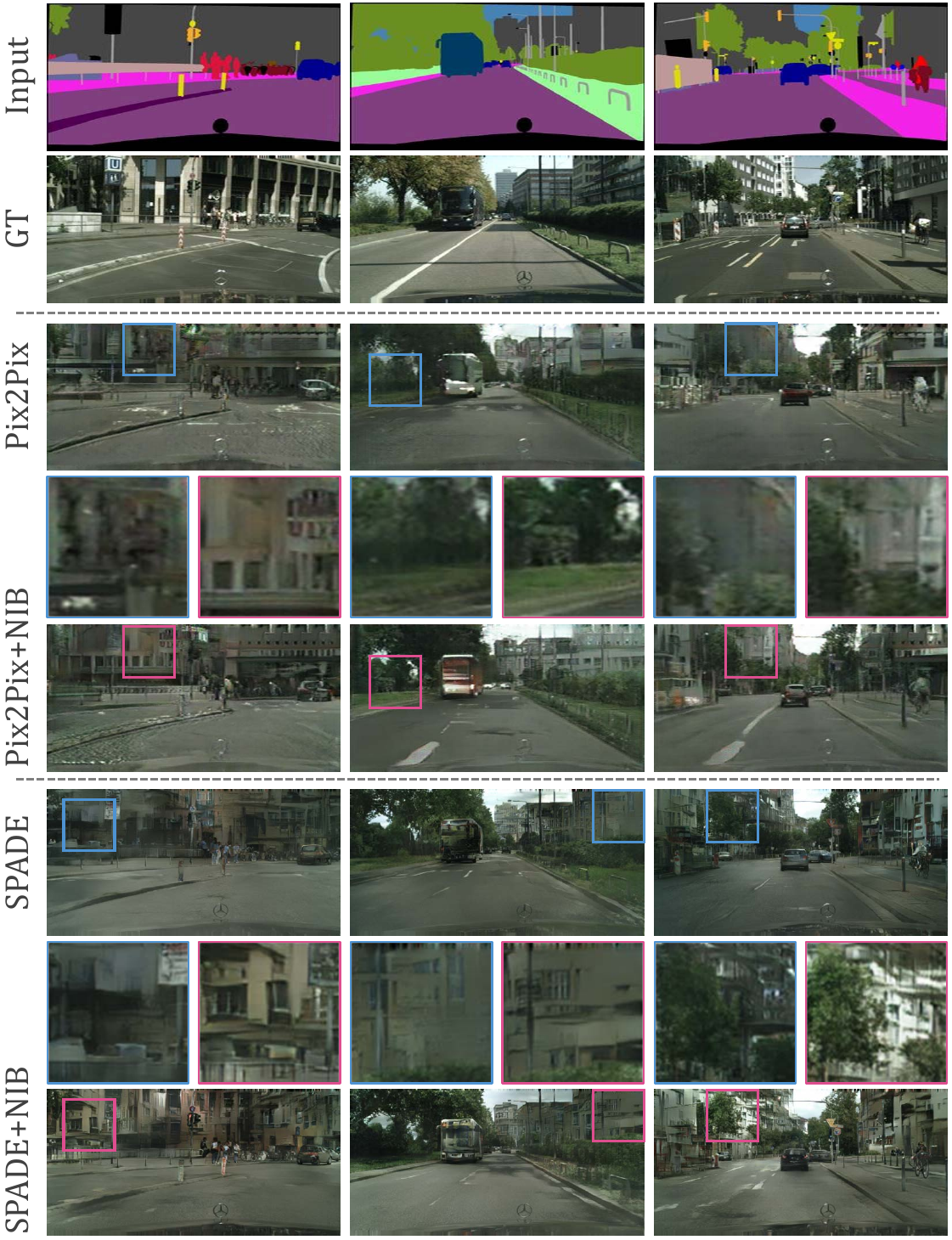}
	\caption{Semantic image synthesis by state-of-the-art models and their NIB-equipped variants. The blowup regions highlight the performance differences.}
	\label{fig:semantic_synthesis}
\end{figure}

Semantic image synthesis aims to synthesize photorealistic images from semantic layout. In general, the input semantic layout consists of constant-valued segments. According to our analysis, such flat input may hinder the CNN performance with flatness degradation, which motivates us to use NIB in this task. Particularly, we investigate the benefits by equipping NIBs into two typical semantic synthesis models, Pix2Pix~\cite{IsolaCVPR17} and SPADE~\cite{ParkCVPR19}.
Figure~\ref{fig:semantic_synthesis} compares the results on several typical examples. Comparing to the original variants, the NIB-equipped models exhibit advantages in synthesizing more realistic textures with sharp details.
Although the CNN has a receptive field covering the whole input image, the shallow convolution layers still degrades in local flatness areas, which affects the model flexibility in some degree. In contrast, the NIB-equipped model is free of flatness degradation and enables the CNN to work with full capability.

\begin{table}[!t]
	\centering
	\small
	\renewcommand{\tabcolsep}{12pt}
	\renewcommand\arraystretch{1.2}
	\caption{Quantitative evaluation on semantic image synthesis on \textit{Cityscapes} dataset.}
	\begin{tabular}{cccc}
		\hline
		Method         & mIOU $\uparrow$  & Accu $\uparrow$   & FID $\downarrow$   \\ \hline
	    Pix2Pix~\shortcite{IsolaCVPR17}           & 12.05\%	  	  & 53.6\%           & 92.9               \\
	    Pix2Pix+NIB        & \bf{12.27\%}   		  & 53.6\%           & \bf{88.1}               \\ \hline\hline
	    SPADE~\shortcite{ParkCVPR19}          & \bf{62.3}\%	  	  & \bf{81.9}\%           & 71.8               \\
	    SPADE+NIB      & 61.9\%   		  & 81.8\%           & \bf{54.6}               \\ \hline
	\end{tabular}
	\label{tab:semantic_synthesis} 
\end{table}

We perform quantitative evaluation on the testing set of~\textit{Cityscapes}~\cite{Cordts2016}.
A pretrained FCN model~\cite{Yu17} is utilized to perform segmentation on the synthesized images. The segmentation performance is measured by mean Intersection-over-Union (mIOU) and pixel accuracy, which reflect how the synthesized images are aligned with the corresponding input semantic layouts. Besides, we further calculate the distribution distances between the generated images and real images by the Fréchet Inception Distance (FID)~\cite{HeuselNIPS17}. The results tabulated in Table~\ref{tab:semantic_synthesis} show that our NIB makes notably improvement in FID while no gain in the indirect segmentation accuracy.  These results are consistent with the properties of NIB as it benefits spatially variant generation while having negligible influences on pixel alignment.

\subsection{Data-Hidden Image Generation}
\label{subsec:image_encoding}

\begin{figure}[!t]
	\centering
	\includegraphics[width=1.0\linewidth]{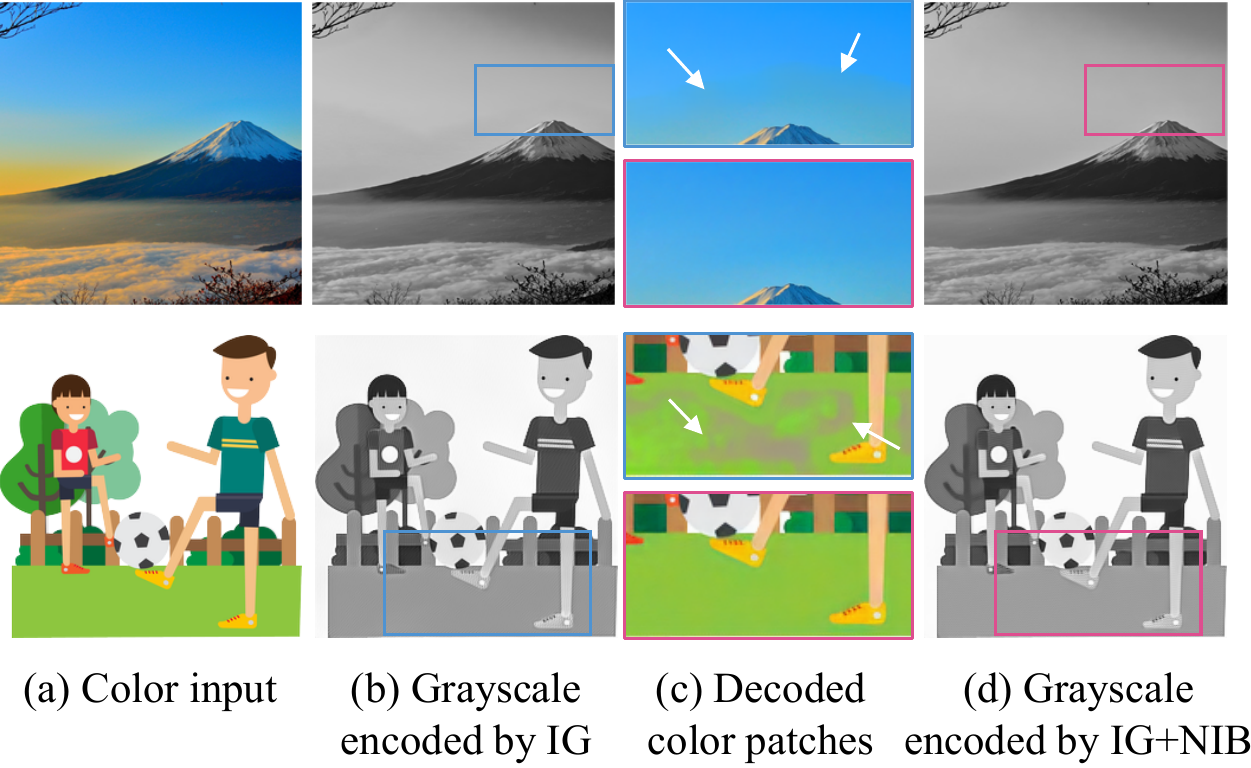}\vspace{-0.5em}
	\caption{Color-hidden grayscale generated by IG model~\protect\shortcite{XiaL2018} and NIB-equipped IG model. Due to flatness degradation, the color-encoded textures can not be generated by IG model in flat areas and color restoration fails consequently. White arrows point to errors.}
	\label{fig:color_encoding}\vspace{-1.0em}
\end{figure}

Invertible grayscale (IG)~\cite{XiaL2018} refers to a kind of grayscale image that has the original color information represented as certain imperceptible texture patterns. Based on the learned encoding scheme via CNN model, it works very well in all kinds of cases except for constant patches. Figure~\ref{fig:color_encoding}(b)(c) shows two examples, where the original colors fail to be recovered from the invertible grayscales. To spot the cause, readers are recommended to zoom in the grayscale images and carefully inspect the weak texture patterns. It shows that no color-encoded textures are generated in those flat or smooth regions, due to flatness degradation. In contrast, with the NIB plugged to~\cite{XiaL2018}, these failure cases are addressed successfully, as the results compared in Figure~\ref{fig:color_encoding}.
Table~\ref{tab:color_encoding} tabulates the quantitative evaluation on the testing set of \textit{DIV2K} dataset~\cite{Agustsson17} (resized to $1024 \times 1024$). The statistic result shows less significant advantages because natural images seldom cover pure flatness. So, the advantages of the NIB-equipped models can be summarized two-fold: it can handles extreme cases (e.g. commic pictures) well; it also benefit general cases to some extent.
More qualitative results are available in the supplementary material.

\begin{table}[!t]
	\centering
	\small
	\renewcommand{\tabcolsep}{15pt}
	\renewcommand\arraystretch{1.2}
	\caption{Quantitative evaluation on data-hidden image generation. The color restoration is performed on \textit{DIV2K} dataset.}\vspace{-0.5em}
	\begin{tabular}{ccc}
		\hline
		Method           & PSRN $\uparrow$    & SSIM $\uparrow$ 	  	  \\ \hline
	    Color-hiding~\shortcite{XiaL2018}               & 38.411	          & 0.9765      \\
	    Color-hiding+NIB           & \textbf{39.314}             & \textbf{0.9811}              \\ \hline
	\end{tabular}
	\label{tab:color_encoding}
\end{table}

\subsection{Deep Neural Dithering}
\label{subsec:fast_halftoning}

\begin{figure}[!t]
	\centering
	\includegraphics[width=1.0\linewidth]{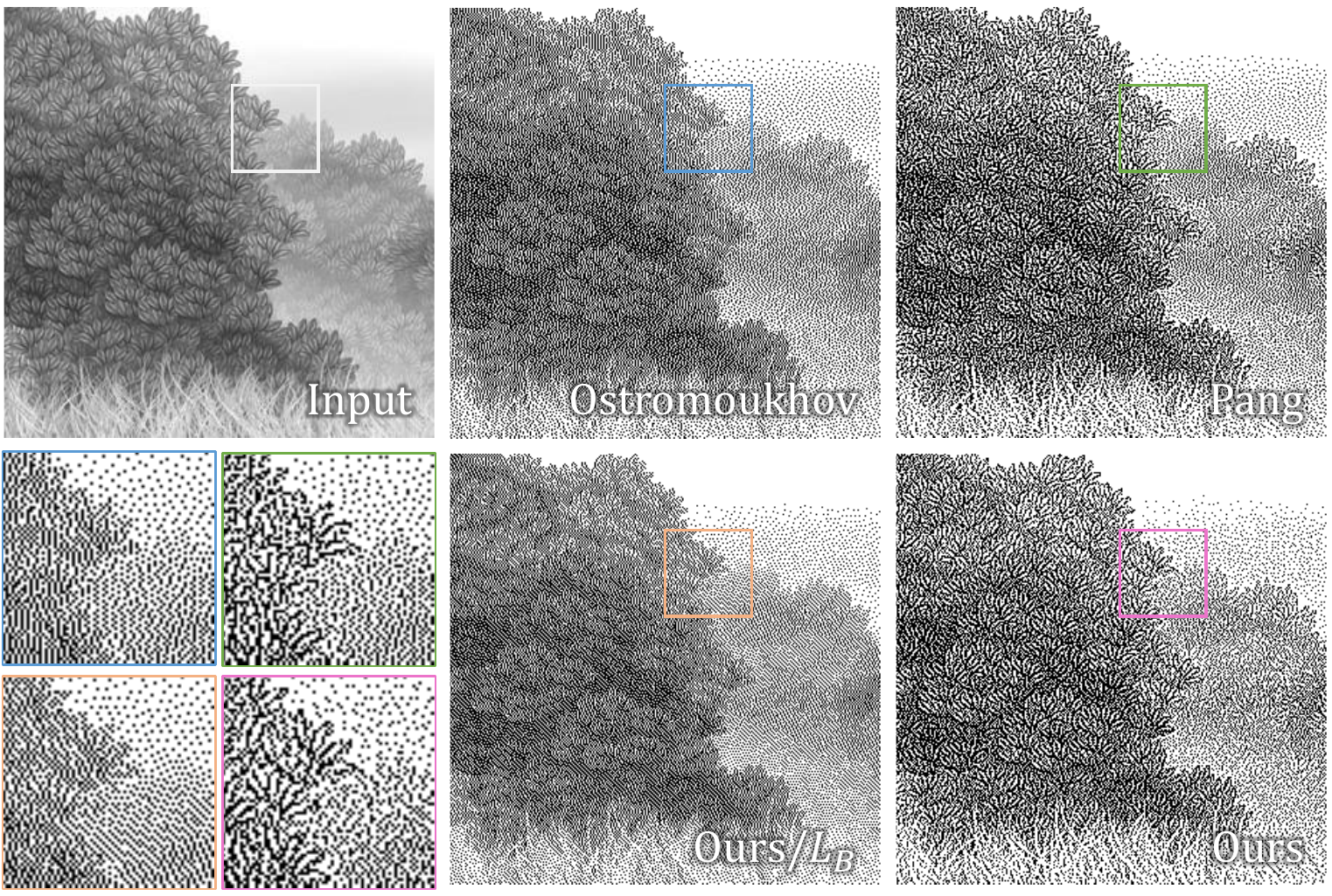}\vspace{-0.5em}
	\caption{Image dithering comparison among Ostromoukhov et al.~\protect\shortcite{Ostromoukhov2001}, Pang et al.~\protect\shortcite{PangQ2008}, our dithering model without blue-noise loss $\mathcal{L}_B$ and our NIB-equipped dithering model. Blowups offer better inspection on halftone patterns.}
	\label{fig:image_dithering}\vspace{-1.0em}
\end{figure}

Early dithering methods mainly focus on tone simulation and blue-noise profile~\cite{Ostromoukhov2001}, and generally suffer from blurry structures. Pang et al.~\cite{PangQ2008} propose an optimization based method that balances clear structures and good blue-noise profile. However, this method is time-consuming and sensitive to parameter tuning, which makes it less practical in applications. To tackle these problems, we make the first attempt to apply neural networks for image dithering. By taking the same loss function of~\cite{PangQ2008}, the CNN model can generates halftone images with good tone similarity (measured by PSNR) but fails to achieve desired blue-noise profile (see those visually annoying strip pattens in the orange blowup in Figure~\ref{fig:image_dithering}). Since blue noise depicts a kind of point distribution without high-frequency component, we formulate an additional loss term to encourage this target: $\mathcal{L}_B=||(\text{DCT}(\mathbf{H})\odot\mathbf{M}||_1$, where DCT($\cdot$) denotes the discrete cosine transformation (DCT), $\odot$ means Hadamard product, and $\mathbf{M}$ denotes a
constant binary mask with $\%5$ low-frequency components set to 0 and others set to 1 in frequency domain. Considering the low-frequency assumption, this constraint can only be imposed on the dithered image $\mathbf{H}$ of constant-grayness, which challenges standard CNNs but is well enabled by NIB-equipped CNNs.
Figure~\ref{fig:image_dithering} shows the comparative results, which evidences the effectiveness of our proposed deep neural dithering method. Also, the quantitative comparison on the classic~\textit{Structure} dataset~\cite{PangQ2008} is tabulated in Table~\ref{tab:neural_dithering}.
Note that, besides of the better quality, our proposed neural dithering is parameter-free and as efficient as a CNN forward process.

\begin{table}[!t]
	\centering
	\small
	\renewcommand{\tabcolsep}{9pt}
	\renewcommand\arraystretch{1.2}
	\caption{Quantitative evaluation on image dithering on \textit{Structure} dataset.}\vspace{-0.5em}
	\begin{tabular}{ccc}
		\hline
		Method           			   & PSRN $\uparrow$    & SSIM $\uparrow$ 	  	  \\ \hline
		Ostromoukhov halftoning~\shortcite{Ostromoukhov2001}     & \textbf{39.935}	            & 0.2425               \\
		Pang halftoning~\shortcite{PangQ2008}     & 34.987	            & 0.3032               \\
		Deep neural dithering/$\mathcal{L}_B$  & 37.392             & 0.2728               \\
	    Deep neural dithering           & 36.434             & \textbf{0.3078}              \\ \hline
	\end{tabular}
	\label{tab:neural_dithering}
\end{table}

\section{Conclusion}
\label{sec:conclusion}

We illustrate the flatness degradation concept of CNNs through typical applications, and proposed a generic solution to address it. It is a simple but effective convolutional plug-in, Noise Incentive Block (NIB), which adaptively turns input data into non-flat features while reserves high fidelity. The rationale and effectiveness of this design are studied in-depth. Extensive experiments show that NIB-equipped CNN models are free of flatness degradation, and thus can handle previously intractable cases. As micro-flatness commonly exists, statistical results evidence the advantage of NIB in general cases. All these results justify the importance of our proposed NIB, as a model-agnostic plug-in for CNNs.
Additionally, the concept presented in this paper is expected to inspire further studies in this direction.

\medskip
\small
\bibliographystyle{ijcai21}
\bibliography{symnoise}

\begin{thebibliography}{}

\bibitem[\protect\citeauthoryear{Agustsson and Timofte}{2017}]{Agustsson17}
Eirikur Agustsson and Radu Timofte.
\newblock Ntire 2017 challenge on single image super-resolution: Dataset and
  study.
\newblock In {\em IEEE Conference on Computer Vision and Pattern Recognition
  Workshops (CVPRW)}, 2017.

\bibitem[\protect\citeauthoryear{Barrera and Jara}{2015}]{Gerardo2015}
Gerardo Barrera and Milton Jara.
\newblock Thermalisation for stochastic small random perturbations of
  hyperbolic dynamical systems.
\newblock {\em arXiv preprint:1510.09207}, 2015.

\bibitem[\protect\citeauthoryear{Bottou}{2010}]{Leon17}
Léon Bottou.
\newblock Large-scale machine learning with stochastic gradient descent.
\newblock In {\em Proceedings of COMPSTAT}, 2010.

\bibitem[\protect\citeauthoryear{Cordts \bgroup \em et al.\egroup
  }{2016}]{Cordts2016}
Marius Cordts, Mohamed Omran, Sebastian Ramos, Timo Rehfeld, Markus Enzweiler,
  Rodrigo Benenson, Uwe Franke, Stefan Roth, and Bernt Schiele.
\newblock The cityscapes dataset for semantic urban scene understanding.
\newblock In {\em {IEEE} Conference on Computer Vision and Pattern Recognition
  (CVPR)}, 2016.

\bibitem[\protect\citeauthoryear{Dai \bgroup \em et al.\egroup }{2017}]{Dai17}
Jifeng Dai, Haozhi Qi, Yuwen Xiong, Yi~Li, Guodong Zhang, Han Hu, and Yichen
  Wei.
\newblock Deformable convolutional networks.
\newblock In {\em {IEEE} International Conference on Computer Vision (ICCV)},
  2017.

\bibitem[\protect\citeauthoryear{Everingham \bgroup \em et al.\egroup
  }{2012}]{VOC2012}
Mark Everingham, SM~Ali Eslami, Luc~Van Gool, Christopher~KI Williams, John
  Winn, and Andrew Zisserman.
\newblock The pascal visual object classes challenge 2012 (voc2012) results.
\newblock
  http://www.pascal-network.org/challenges/VOC/voc2012/workshop/index.html,
  2012.

\bibitem[\protect\citeauthoryear{Goodfellow \bgroup \em et al.\egroup
  }{2014}]{GoodfellowNIPS14}
Ian~J. Goodfellow, Jean Pouget-Abadie, Mehdi Mirza, Bing Xu, David
  Warde-Farley, Sherjil Ozair, Aaron Courville, and Yoshua Bengio.
\newblock Generative adversarial nets.
\newblock In {\em Advances in Neural Information Processing Systems (NIPS)},
  2014.

\bibitem[\protect\citeauthoryear{He \bgroup \em et al.\egroup
  }{2016}]{HeCVPR16}
Kaiming He, Xiangyu Zhang, Shaoqing Ren, and Jian Sun.
\newblock Deep residual learning for image recognition.
\newblock In {\em {IEEE} Conference on Computer Vision and Pattern Recognition
  (CVPR)}, 2016.

\bibitem[\protect\citeauthoryear{Hu \bgroup \em et al.\egroup }{2019}]{Hux19}
Xinghong Hu, Xueting Liu, Zhuming Zhang, Menghan Xia, Chengze Li, and
  Tien{-}Tsin Wong.
\newblock Colorblind-shareable videos by synthesizing temporal-coherent
  polynomial coefficients.
\newblock {\em {ACM} Transactions o. Graphics (TOG)}, 38(6):174:1--174:12,
  2019.

\bibitem[\protect\citeauthoryear{Isola \bgroup \em et al.\egroup
  }{2017}]{IsolaCVPR17}
Phillip Isola, Jun{-}Yan Zhu, Tinghui Zhou, and Alexei~A. Efros.
\newblock Image-to-image translation with conditional adversarial networks.
\newblock In {\em {IEEE} Conference on Computer Vision and Pattern Recognition
  (CVPR)}. {IEEE} Computer Society, 2017.

\bibitem[\protect\citeauthoryear{Jin \bgroup \em et al.\egroup }{2015}]{Jin15}
Jonghoon Jin, Aysegul Dundar, and Eugenio Culurciello.
\newblock Robust convolutional neural networks under adversarial noise.
\newblock {\em arXiv preprint:1511.06807}, 2015.

\bibitem[\protect\citeauthoryear{Karras \bgroup \em et al.\egroup
  }{2019}]{KarrasCVPR19}
Tero Karras, Samuli Laine, and Timo Aila.
\newblock A style-based generator architecture for generative adversarial
  networks.
\newblock In {\em {IEEE} Conference on Computer Vision and Pattern Recognition
  (CVPR)}, 2019.

\bibitem[\protect\citeauthoryear{Keskar \bgroup \em et al.\egroup
  }{2017}]{KeskarICLR17}
Nitish~Shirish Keskar, Dheevatsa Mudigere, Jorge Nocedal, Mikhail Smelyanskiy,
  and Ping Tak~Peter Tang.
\newblock On large-batch training for deep learning: Generalization gap and
  sharp minima.
\newblock In {\em International Conference on Learning Representations (ICLR)},
  2017.

\bibitem[\protect\citeauthoryear{Kingma and Welling}{2019}]{Kingma19}
Diederik~P. Kingma and Max Welling.
\newblock An introduction to variational autoencoders.
\newblock {\em Foundations and Trends in Machine Learning}, 12(4):307--392,
  2019.

\bibitem[\protect\citeauthoryear{Martin~Heusel and
  Hochreiter}{2017}]{HeuselNIPS17}
Thomas Unterthiner Bernhard~Nessler Martin~Heusel, Hubert~Ramsauer and Sepp
  Hochreiter.
\newblock Gans trained by a two time-scale update rule converge to a local nash
  equilibrium.
\newblock In {\em Advances in Neural Information Processing Systems (NIPS)},
  2017.

\bibitem[\protect\citeauthoryear{Neelakantan \bgroup \em et al.\egroup
  }{2015}]{Neelakantan2015}
Arvind Neelakantan, Luke Vilnis, Quoc~V. Le, Ilya Sutskever, Lukasz Kaiser,
  Karol Kurach, and James Martens.
\newblock Adding gradient noise improves learning for very deep networks.
\newblock {\em arXiv preprint:1511.06807}, 2015.

\bibitem[\protect\citeauthoryear{Noh \bgroup \em et al.\egroup
  }{2017}]{NohNIPS17}
Hyeonwoo Noh, Tackgeun You, Jonghwan Mun, and Bohyung Han.
\newblock Regularizing deep neural networks by noise: Its interpretation and
  optimization.
\newblock In {\em Advances in Neural Information Processing Systems (NIPS)},
  2017.

\bibitem[\protect\citeauthoryear{Ostromoukhov}{2001}]{Ostromoukhov2001}
Victor Ostromoukhov.
\newblock A simple and efficient error-diffusion algorithm.
\newblock In {\em ACM SIGGRAPH}, 2001.

\bibitem[\protect\citeauthoryear{Pang \bgroup \em et al.\egroup
  }{2008}]{PangQ2008}
Wai{-}Man Pang, Yingge Qu, Tien{-}Tsin Wong, Daniel Cohen{-}Or, and Pheng{-}Ann
  Heng.
\newblock Mononizing binocular videos.
\newblock In {\em SIGGRAPH}, 2008.

\bibitem[\protect\citeauthoryear{Park \bgroup \em et al.\egroup
  }{2019}]{ParkCVPR19}
Taesung Park, Ming{-}Yu Liu, Ting{-}Chun Wang, and Jun{-}Yan Zhu.
\newblock Semantic image synthesis with spatially-adaptive normalization.
\newblock In {\em {IEEE} Conference on Computer Vision and Pattern Recognition
  (CVPR)}, 2019.

\bibitem[\protect\citeauthoryear{Simonyan and Zisserman}{}]{Simonyan2015}
Karen Simonyan and Andrew Zisserman.
\newblock Very deep convolutional networks for large-scale image recognition.
\newblock In {\em International Conference on Learning Representations (ICLR),
  year = {2015}}.

\bibitem[\protect\citeauthoryear{Su \bgroup \em et al.\egroup }{2019}]{SuH19}
Hang Su, Varun Jampani, Deqing Sun, Orazio Gallo, Erik~G. Learned{-}Miller, and
  Jan Kautz.
\newblock Pixel-adaptive convolutional neural networks.
\newblock In {\em {IEEE} Conference on Computer Vision and Pattern Recognition
  (CVPR)}, 2019.

\bibitem[\protect\citeauthoryear{Ulichney}{1987}]{Ulichney1987}
Robert Ulichney.
\newblock {\em Digital halftoning}.
\newblock MIT press, 1987.

\bibitem[\protect\citeauthoryear{Verma \bgroup \em et al.\egroup
  }{2016}]{Verma17}
Nitika Verma, Edmond Boyer, and Jakob Verbeek.
\newblock Dynamic filters in graph convolutional networks.
\newblock 2016.

\bibitem[\protect\citeauthoryear{Xia \bgroup \em et al.\egroup
  }{2018}]{XiaL2018}
Menghan Xia, Xueting Liu, and Tien-Tsin Wong.
\newblock Invertible grayscale.
\newblock {\em {ACM} Transactions on Graphics (TOG)}, 37(6):246:1--246:10,
  2018.

\bibitem[\protect\citeauthoryear{Yu \bgroup \em et al.\egroup }{2017}]{Yu17}
Fisher Yu, Vladlen Koltun, and Thomas~A. Funkhouser.
\newblock Dilated residual networks.
\newblock In {\em {IEEE} Conference on Computer Vision and Pattern Recognition
  (CVPR)}, 2017.

\bibitem[\protect\citeauthoryear{Zhang \bgroup \em et al.\egroup
  }{2017}]{ZhangICLR17}
Chiyuan Zhang, Samy Bengio, Moritz Hardt, Benjamin Recht, and Oriol Vinyals.
\newblock Understanding deep learning requires rethinking generalization.
\newblock In {\em International Conference on Learning Representations (ICLR)},
  2017.

\bibitem[\protect\citeauthoryear{Zhou \bgroup \em et al.\egroup
  }{2019}]{ZhouICML2019}
Mo~Zhou, Tianyi Liu, Yan Li, Dachao Lin, Enlu Zhou, and Tuo Zhao.
\newblock Towards understanding the importance of noise in training neural
  networks.
\newblock 2019.

\bibitem[\protect\citeauthoryear{Zhu \bgroup \em et al.\egroup
  }{2017}]{ZhuNIPS17}
Jun{-}Yan Zhu, Richard Zhang, Deepak Pathak, Trevor Darrell, Alexei~A. Efros,
  Oliver Wang, and Eli Shechtman.
\newblock Toward multimodal image-to-image translation.
\newblock In {\em Advances in Neural Information Processing Systems (NIPS)},
  2017.

\end{thebibliography}

\end{document}